\documentclass[conference]{IEEEtran}
\IEEEoverridecommandlockouts

\usepackage{cite}
\usepackage{amsmath,amssymb,amsfonts}
\usepackage{algorithmic}
\usepackage{graphicx}
\usepackage{textcomp}
\usepackage{xcolor}
\usepackage{url}

\usepackage{multirow}

\begin{document}

\title{KANLib - A Modular, Extensible and Fast Kolmogorov-Arnold Network Implementation}

\author{\IEEEauthorblockN{1\textsuperscript{st} Julian Hoever}
\IEEEauthorblockA{\textit{Intelligent Embedded Systems} \\
\textit{University of Duisburg-Essen}\\
Duisburg, Germany \\
julian.hoever@uni-due.de}
\and
\IEEEauthorblockN{2\textsuperscript{nd} Gregor Schiele}
\IEEEauthorblockA{\textit{Intelligent Embedded Systems} \\
\textit{University of Duisburg-Essen}\\
Duisburg, Germany \\
gregor.schiele@uni-due.de}
}

\maketitle

\begin{abstract}
Kolmogorov-Arnold Networks (KANs) have recently emerged as a promising alternative to traditional multilayer perceptrons by replacing linear weights with learnable univariate functions. Despite their theoretical advantages in interpretability and expressiveness, practical research of KANs remains difficult due to high computational costs and inconsistent feature support across existing frameworks. This paper introduces \textit{KANLib}, a modular, extensible, and computationally efficient framework for developing and evaluating KAN architectures. KANLib unifies core concepts from existing implementations, including PyKAN, EfficientKAN, and FastKAN, within a consistent software architecture that emphasizes flexibility, feature parity, and high performance. The framework supports two basis function types, adaptive grid rescaling, grid extension, and fine-grained architectural customization while maintaining compatibility with standard PyTorch workflows. Experimental evaluation on the California Housing benchmark demonstrates that KANLib reproduces the predictive behavior of established reference KAN implementations while achieving competitive computational efficiency. Furthermore, the framework enables the exploration of architectural variations beyond standard KAN formulations with only minor impacts on predictive performance. Overall, KANLib provides a robust foundation for future research on scalable and extensible KAN architectures.
\end{abstract}


\section{Introduction}\label{sec:introduction}

Artificial neural networks have become the dominant paradigm in modern machine learning, achieving remarkable performance across diverse domains such as computer vision, natural language processing, and audio analysis. Despite the variety of architectures developed for these fields, most share a common structural principle: information is processed through a sequence of affine transformations followed by a fixed non-linear activation function. Formally, each layer computes $y = \sigma(Wx + b)$ where $W$ and $b$ are learnable parameters, and $\sigma$ denotes a non-linear activation function (e.g. ReLU).

In traditional multilayer perceptrons (MLPs), this framework has been shown to possess universal approximation capabilities when properly configured and trained \cite{hornik_multilayer_1989}. A key limitation of MLPs lies in their reliance on a fixed activation function: only the affine parameters $W$ and $b$ are trainable. This constraint limits interpretability, as each neuron’s output is influenced by all linearly transformed inputs from the previous layer. As a result, individual non-linear relationships between inputs and outputs within a single MLP layer cannot be captured directly.

To enhance interpretability, particularly in scientific discovery tasks, Liu et al. \cite{liu_kan_2025} proposed Kolmogorov-Arnold networks (KANs), a novel framework to construct multi-layer networks in which the linear parameters of MLPs are replaced by learnable non-linear functions. Formally, the $j$-th output of a single KAN layer is defined as $y_j = \sum_i \Phi_{i,j}(x_i)$ where $\Phi_{i,j}$ are learnable non-linear functions. This design eliminates the need for fixed activation functions, as each connection in the network inherently introduces its own non-linearity.

Despite their potential, the evaluation of KANs in practical applications is often hindered by high computational cost, complex implementations, or a lack of consistent features across different existing implementations. To bridge the gap between KAN theory and practical application, this work introduces an optimized framework, KANLib \cite{hoever_julianhoeverkanlib_2026}, designed with the following objectives in mind:
\begin{itemize}
	\item \textbf{Computational Efficiency}: The framework should support fast training and inference to make KANs feasible for larger datasets.
	\item \textbf{Modularity and Extensibility}: The framework should be modular and allow to easily add new features in order to accelerate the development of new KAN experiments.
	\item \textbf{Consistent Feature Set}: The framework should facilitate rigorous benchmarking by providing a consistent suite of features across all supported KAN implementations where applicable.
\end{itemize}

The remainder of this paper is structured as follows: 
Section~\ref{sec:kan_basics} introduces the theoretical concept of KANs, while Section~\ref{sec:kan_implementations} describes three established implementations.
Section~\ref{sec:kanlib} presents our KANLib framework combining contributions from the previously described implementations.
Section~\ref{sec:evaluation} evaluates KANLib's predictive performance and computational efficiency. 
Finally, Section~\ref{sec:conclusion} concludes the paper and outlines future work.

\section{Kolmogorov-Arnold Networks}\label{sec:kan_basics}

A fundamental concept underlying KANs is the Kolmogorov-Arnold representation theorem \cite{kolmogorov_representation_1957}. The theorem states that each continuous multivariate function $f: [0,1]^n \to \mathbb{R}$ with $n \geq 2$ can be represented as finite sums of univariate functions $\psi_{p,q}: [0,1] \to \mathbb{R}$ and $\chi_{q}: \mathbb{R} \to \mathbb{R}$ such that
\begin{equation}
	f(x_1, \dots, x_n) = \sum_{q=1}^{2n+1} \chi_q(\sum_{p=1}^n \psi_{p,q}(x_p))
	\label{eq:kolmogorov_arnold_theorem}
\end{equation}

Equation~\ref{eq:kolmogorov_arnold_theorem} naturally suggests the structure of a two-layer network with the input features $x_1, \dots, x_n$, the networks output $f$ and $\psi_{p,q}$, $\chi_q$ as learnable functions of the first and second layer. However, there is some criticism that the theorem cannot be applied to machine learning in practice, due to the potentially non-smooth behavior of the inner functions $\psi_{p,q}$ \cite{girosi_representation_1989}.

Liu et al. \cite{liu_kan_2025} take a more optimistic perspective on the usability of the Kolmogorov-Arnold representation theorem in machine learning and generalize Equation ~\ref{eq:kolmogorov_arnold_theorem} to an arbitrary number of layers (depth) and output features (width).

To formalize this, let $L$ denote the number of KAN layers, and let $n_1, \dots n_L$ represent the number of output features for each layer. The number of input features in the first layer is denoted by $n_0$. Each KAN layer $l \in \{1, \dots, L\}$ is associated with a matrix $\Phi^{(l)}$ of shape $n_l \times n_{l-1}$, whose entries are learnable univariate functions \cite{liu_kan_2025}.

Using this, the output feature vector $x^{(l)} \in \mathbb{R}^{n_l}$ of a layer for a given input feature vector $x^{(l-1)} \in \mathbb{R}^{n_{l-1}}$ can be computed as follows \cite{liu_kan_2025}:

\begin{equation}
	x^{(l)} = \Phi^{(l)} \circ x^{(l-1)} = 
	\begin{pmatrix}
		\sum_{i=1}^{n_{l-1}} \phi_{1,i}^{(l)}(x_i^{(l-1)})  \\
		\vdots \\
		\sum_{i=1}^{n_{l-1}} \phi_{n_l,i}^{(l)}(x_i^{(l-1)}) 
	\end{pmatrix}
\end{equation}

Therefore, a KAN network can be defined by successively applying the input vector $x \in \mathbb{R}^{n_0}$ to the matrices of learnable univariate functions:

\begin{equation}
	\text{KAN}(x) = \Phi^{(L)} \circ \dots (\Phi^{(2)} \circ (\Phi^{(1)} \circ x))\dots
\end{equation}

In a KAN, each learnable univariate function $\phi$ can model a non-linear relationship, allowing each layer to directly represent non-linear influences of individual input features on (intermediate) output features. In practice, learnable univariate functions are typically realized as splines \cite{liu_kan_2025,li_kolmogorov-arnold_2024,blealtan_blealtanefficient-kan_2025}.

\section{KAN Implementations}\label{sec:kan_implementations}

Since the publication of the KAN architecture by Liu et al. \cite{liu_kan_2025} in 2024, the field has seen a significant increase in research work. While many approaches utilize spline-based architectures \cite{liu_kan_2025,li_kolmogorov-arnold_2024,blealtan_blealtanefficient-kan_2025}, others explore alternative methods for realizing learnable univariate functions, such as Fourier series \cite{xu_enhancing_2025} or Jacobi basis functions \cite{afzal_aghaei_fkan_2025}.

This section focuses on three prominent spline-based KAN implementations. Specifically, the official PyKAN framework of Liu et al. \cite{liu_kan_2025,liu_kindxiaomingpykan_2025} is evaluated alongside two computationally efficient versions: EfficientKAN \cite{blealtan_blealtanefficient-kan_2025} and FastKAN \cite{li_kolmogorov-arnold_2024,li_ziyaolifast-kan_2025}. Subsequently, in Section~\ref{sec:kanlib}, the KANLib framework \cite{hoever_julianhoeverkanlib_2026} is introduced, which consolidates features from these implementations to provide a fast and modular environment for KAN research. All the considered implementations support MLP-like KAN models only (no convolutional KANs, etc.).

The PyKAN, EfficientKAN, and FastKAN frameworks were selected because they are widely recognized and provide a consistent baseline for research. While numerous other KAN variants have emerged in recent years \cite{jinhuilin_mintisanawesome-kan_2026}, they remain outside the scope of this comparison.

\subsection{PyKAN}

The seminal work by Liu et al. \cite{liu_kan_2025} generalized the Kolmogorov-Arnold representation theorem to networks of arbitrary depth, as discussed in Section~\ref{sec:kan_basics}. The official implementation, PyKAN \cite{liu_kindxiaomingpykan_2025}, utilizes B-splines to model the univariate continuous non-linear functions $\phi$ within the weight matrix. To enhance optimizability, the authors define each learnable function $\phi$ as a weighted combination of a learnable B-spline $\textit{spline}(x)$ and a residual branch $b(x)$ \cite{liu_kan_2025}:

\begin{equation}
	\phi(x) = w_b b(x) + w_s \text{spline}(x)
	\label{eq:learnable_function}
\end{equation}

In this definition, the residual branch $b(x)$ employs the SiLU (Sigmoid Linear Unit) activation function, $b(x) = \text{SiLU}(x) = x / (1 + \exp(-x))$. The spline component is defined as $\textit{spline}(x) = \sum_i c_i B_i(x)$ with the  B-spline basis functions $B_i$ and their corresponding learnable coefficients $c_i$.

Beyond the core architecture, PyKAN \cite{liu_kindxiaomingpykan_2025} provides an extensive suite of functionality for model creation, manipulation and training. The PyKAN framework is designed to be an interactive scientific discovery tool \cite{liu_kindxiaomingpykan_2025}. Therefore, a primary focus of the framework is interpretability; it includes built-in functionality for pruning and symbolic regression by replacing learned splines with symbolic mathematical expressions (e.g., $\sin$ or $\exp$). This allows researchers to transform complex neural representations into transparent, closed-form equations.

To facilitate the interpretation of these representations, PyKAN incorporates visualization capabilities. Researchers can render the entire network graph, where each edge displays the shape of its corresponding learned univariate function $\phi$.

Furthermore, the framework supports sophisticated grid manipulation techniques to refine the accuracy of the spline approximations during the training process.

\subsubsection{Grid Manipulation}\label{sec:pykan_grid_manipulation}

A significant advantage of KANs is that their learnable edges are continuous univariate functions, which can be dynamically re-projected onto different grids without losing learned information. PyKAN leverages this property through two main mechanisms:

\begin{itemize}
	\item \textbf{Grid Extension}: When creating a KAN, an initial grid size has to be specified. Grid extension allows to gradually learn higher-frequency details during training by replacing a coarse-grained grid with a finer-grained grid and projecting the already-learned spline onto the fine-grained grid to make the spline more accurate \cite{liu_kan_2025}.
	\item \textbf{Adaptive Grid Rescaling}: During training, input values of layers in KANs may shift outside the initial grid range or input values are not equally distributed over all grid points. PyKAN addresses this by updating the grid boundaries and distribution based on the input data statistics.
\end{itemize}

\subsection{EfficientKAN}\label{sec:efficient_kan}

The initial PyKAN implementation faced performance issues due to the high computational overhead of B-spline basis function evaluations. Furthermore, to support the extensive feature set offered by PyKAN, the implementation computes and composes various intermediate results. This architectural choice limits the ability to represent large segments of the computational graph using highly efficient PyTorch operations. To address the computational limitations of PyKAN, EfficientKAN \cite{blealtan_blealtanefficient-kan_2025} introduces three main optimizations: a more efficient B-spline basis evaluation, an optimized KAN linear layer computation, and a faster residual branch implementation.

The original PyKAN framework expands the input tensor from a shape of $(n_\text{batch}, n_\text{inputs})$ to $(n_\text{batch}, n_\text{outputs}, n_\text{inputs})$, duplicating inputs for each output feature. Since B-spline basis evaluation is computationally expensive, this increases the evaluation cost by a factor of $n_\text{outputs}$. EfficientKAN avoids this expansion by computing B-spline basis values directly from the input. This optimization was later integrated into PyKAN in June 2024 \cite{liu_kan_2024}.

EfficientKAN further combines the B-spline basis weighting and linear layer computation into a single, highly optimized \verb|torch.nn.functional.linear| function call. The residual branch weighting is implemented similarly.

However, these coarse-grained PyTorch operations omit intermediate activations, making some PyKAN features infeasible, including the activation-based $L_1$ regularization proposed by Liu et al. \cite{liu_kan_2025}. Instead, EfficientKAN applies standard $L_1$ regularization to the spline coefficients \cite{blealtan_blealtanefficient-kan_2025}.

Overall, EfficientKAN prioritizes computational efficiency over feature completeness. The framework supports MLP-like KAN models, adaptive grid rescaling similar to PyKAN, and coefficient-based $L_1$ regularization \cite{blealtan_blealtanefficient-kan_2025}.

\subsection{FastKAN}

The FastKAN framework \cite{li_kolmogorov-arnold_2024, li_ziyaolifast-kan_2025}, developed by Ziyao Li, builds upon the optimizations introduced by EfficientKAN to further accelerate KANs. Its main contribution is to replace third-order B-spline basis functions with Gaussian radial basis functions (RBFs) \cite{li_kolmogorov-arnold_2024}.

\subsubsection{Gaussian Radial Basis Functions}

The primary motivation for utilizing Gaussian RBFs is to circumvent the computational complexity associated with B-splines. A Gaussian radial basis function $B_{\text{RBF},i}$ for an input $x$, a center point $x_i$, and a spread $\sigma$ is defined as:

\begin{equation}
	B_{\text{RBF},i}(x) = \exp \left( -\frac{||x - x_i||^2}{2\sigma^2} \right)
    \label{eq:gaussian_rbf}
\end{equation}

By substituting the B-spline basis $B_i$ in the spline computation, with this Gaussian RBF, FastKAN eliminates the need for the iterative or recursive computations (such as the Cox-deBoor recursion) typically required for B-spline evaluations. This allows for direct computation of the basis function values, promising a significant increase in training and inference throughput.

\subsubsection{Grid Considerations}\label{sec:fastkan_grid_considerations}

In the context of FastKAN, the grid of a spline is defined by the center points $x_i$ of the Gaussian RBFs (see Equation~\ref{eq:gaussian_rbf}). While Gaussian RBFs offer computational advantages, their ability to approximate third-order B-spline basis functions is highly dependent on the grid's distribution. As demonstrated by Ziyao Li, Gaussian RBFs can effectively approximate B-spline basis functions through linear transformations, provided the grid is equidistant \cite{li_kolmogorov-arnold_2024}.

\begin{figure}
	\centering
	\includegraphics[width=\linewidth]{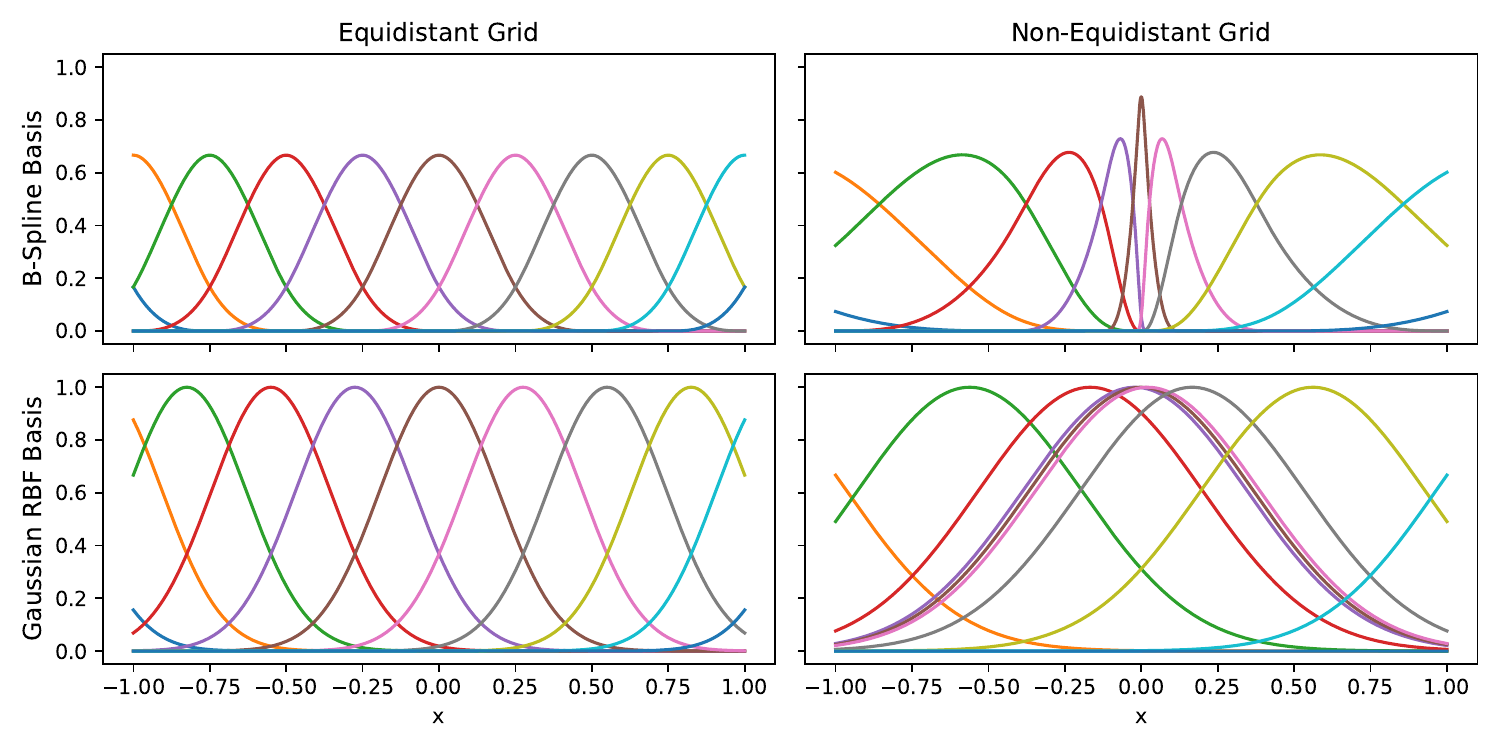}
	\caption{Examples of B-spline and Gaussian radial basis functions with equi- and non-equidistant grids.}
	\label{fig:basis_functions}
\end{figure}

However, this approximation capability breaks down on non-equidistant grids. As illustrated in Figure~\ref{fig:basis_functions}, Gaussian RBFs fail to maintain the necessary local characteristics to mimic B-spline basis function behavior when grid spacing is not equidistant, as they cannot approximate the B-spline basis functions through linear transformations alone in such configurations.

\subsubsection{Implementation}

Consistent with its focus on computational performance, the FastKAN implementation \cite{li_ziyaolifast-kan_2025} offers a more restricted feature set compared to both PyKAN \cite{liu_kindxiaomingpykan_2025} and EfficientKAN \cite{blealtan_blealtanefficient-kan_2025}. The framework primarily supports the construction of MLP-like KAN models and provides basic visualization functionality for plotting learned spline functions.

Notably, FastKAN does not support the adaptive grid rescaling mechanism found in PyKAN and EfficientKAN. To compensate the lack of grid rescaling abilities, it applies Layer Normalization \cite{ba_layer_2016} to the inputs before the basis function evaluation, effectively mapping the input distribution to the RBFs' active range \cite{li_ziyaolifast-kan_2025}.

\section{KANLib}\label{sec:kanlib}

To accelerate the research and development of KANs, our KANLib framework \cite{hoever_julianhoeverkanlib_2026} synthesizes the core innovations of PyKAN \cite{liu_kan_2025,liu_kindxiaomingpykan_2025}, EfficientKAN \cite{blealtan_blealtanefficient-kan_2025}, and FastKAN \cite{li_kolmogorov-arnold_2024,li_ziyaolifast-kan_2025}. The framework is designed to provide researchers with a highly flexible environment for creating and training KAN models while maintaining high computational performance.

The architectural design of KANLib is guided by three primary requirements: computational efficiency, modularity, and feature consistency. While existing implementations provide disparate and often incompatible feature sets, KANLib consolidates these diverse approaches within a unified software architecture. These design goals are specifically tailored to our research needs. By providing a computationally efficient implementation, KANLib allows for the evaluation of KANs on resource-restricted hardware using large datasets.

\subsection{Supported Features}

The modular design ensures that all linear layer types regardless of the underlying basis function, share a consistent feature set. This consistency is critical for facilitating direct, rigorous comparisons between different KAN variations. Currently, KANLib supports the following features:

\begin{itemize}
	\item \textbf{MLP-like KAN models}: Users can create pure MLP-like KAN models by stacking KANLib's linear layers. Furthermore, the framework's design allows these layers to be integrated into any standard PyTorch model to facilitate the development of hybrid architectures.
	\item \textbf{Fine-Grained Control}: Following the definition in Equation~\ref{eq:learnable_function}, KANLib utilizes learnable functions consisting of a spline and a residual branch. However, unlike presented implementations, KANLib provides the granular control to selectively disable the residual branch $w_b b(x)$ or the extra spline weighting $w_s$. Additionally, similar to FastKAN \cite{li_kolmogorov-arnold_2024,li_ziyaolifast-kan_2025}, KANLib supports Layer Normalization \cite{ba_layer_2016} to rescale inputs into the active range of the basis functions.
	\item \textbf{Adaptive Grid Rescaling}: While PyKAN and EfficientKAN offer adaptive grid rescaling for B-splines, FastKAN lacks this mechanism, relying instead on input normalization. KANLib implements the adaptive rescaling strategy for B-spline layers and extends this support to Gaussian RBF-based layers. For Gaussian RBF layers, this is restricted to equidistant grids, as non-equidistant grid configurations are not beneficial (see Section~\ref{sec:fastkan_grid_considerations}).
	\item \textbf{Grid Extension}: A notable feature of PyKAN is the ability to increase spline resolution during training without losing learned information (see Section~\ref{sec:pykan_grid_manipulation}). While omitted in EfficientKAN and FastKAN, KANLib supports grid extension for both B-spline and Gaussian RBF-based linear layers.
	\item \textbf{Training and Visualization}: The framework includes a predefined training function that allows to use adaptive grid rescaling and grid extension steps during training. Additionally, it offers functionality to visualize individual learned spline functions to assist in evaluating the learned knowledge.
\end{itemize}

The KANLib framework is publicly available on GitHub for experimental use \cite{hoever_julianhoeverkanlib_2026}.

\section{Evaluation}\label{sec:evaluation}

The goal of this evaluation is to assess whether KANLib reproduces the predictive behavior and computational characteristics of established KAN implementations while simultaneously providing the architectural flexibility discussed in the previous sections. Rather than focusing exclusively on maximizing predictive accuracy, the experiments are designed to validate the correctness, consistency, and extensibility of the framework across different basis function implementations and layer configurations.

To this end, we compare KANLib against reference implementations presented in Section~\ref{sec:kan_implementations} of both B-spline-based and Gaussian radial basis function (GRBF)-based KAN architectures on the California Housing dataset \cite{noauthor_california_nodate}. The dataset is derived from the 1990 U.S. census and contains 20,640 samples with eight numerical input features, including median income, average number of rooms, and house age. The task is to predict the median house value and therefore provides a simple but well-established regression benchmark for evaluating neural network architectures.

In the following, \textit{KANLib (B-spline)} denotes KANLib models using B-spline basis functions, while \textit{KANLib (GRBF)} refers to KANLib models based on Gaussian radial basis functions.

\subsection{Experimental Setup}\label{sec:experimental_setup}

To ensure a fair and reproducible comparison, all evaluated implementations use the same network architecture and training configuration. Each model consists of a two-layer KAN architecture with a hidden dimension of 30 neurons. Since Gaussian RBFs approximate third-order B-splines, all B-spline-based variants use spline order $3$ in order to maintain architectural consistency across experiments. Furthermore, the grid size is fixed to $10$ and the grid range is defined as $[-1,1]$.

To minimize implementation-specific influences, advanced features such as adaptive grid rescaling and Layer Normalization are intentionally disabled. This allows the comparison to focus on the underlying KAN implementations rather than additional implementation-specific optimization techniques.

The California Housing dataset is split into a training set containing $80\%$ of the samples and a validation set containing the remaining $20\%$. Prior to training, all input features are rescaled to the range $[-1, 1]$ in order to align the input distribution with the B-spline and Gaussian RBF grid domains used in the first layer.

All models are trained for $300$ epochs using the Adam optimizer with a learning rate of $10^{-3}$, a batch size of $1024$, and the mean squared error loss function. To account for stochastic variation caused by random parameter initialization, each experiment is repeated $20$ times using independent weight initialization. The reported metrics correspond to the mean and standard deviation across all runs.

Overall, this evaluation setup provides a standardized environment for comparing different KAN implementations and for validating the consistency and correctness of the KANLib framework.

\subsection{Predictive Performance}

Table~\ref{tbl:baseline} summarizes the predictive performance and computational characteristics of all evaluated implementations on the California Housing dataset. Results are reported as mean values and standard deviations across $20$ independent training runs.

Among the evaluated B-spline-based implementations, KANLib (B-spline) achieves the best predictive performance with an RMSE of $0.5399 \pm 0.0068$ and an coefficient of determination ($R^2$ score) of $0.7834 \pm 0.0054$. Compared to KANLib (B-spline), PyKAN exhibits an approximately $0.2\%$ higher RMSE, while EfficientKAN differs by roughly $1.0\%$. These comparatively small deviations indicate that KANLib successfully reproduces the predictive characteristics of existing spline-based KAN implementations.

A similar observation can be made for the GRBF-based architectures. KANLib (GRBF) achieves an RMSE of $0.5476 \pm 0.0071$ and an $R^2$ score of $0.7772 \pm 0.0058$, while FastKAN differs by only an RMSE that is approximately $0.5\%$ higher. This demonstrates that KANLib is capable of reproducing the behavior of established GRBF-based KAN implementations.

Overall, the predictive results show that KANLib remains fully competitive with widely used reference implementations despite its additional abstraction layers and architectural flexibility.

\begin{table*}
\centering
\caption{Performance on the California Housing dataset across $20$ training runs per implementation. Results are reported as mean $\pm$ standard deviation for validation RMSE, $R^2$, and inference time.}
\begin{tabular}{|l|rr|rr|}
\hline
Implementation    & RMSE                & $R^2$               & $\#_\text{params}$ & Inference Time [ms] \\
\hline
FastKAN           & $0.5501 \pm 0.0052$ & $0.7751 \pm 0.0043$ & $3811$             & $41.15  \pm 0.52$   \\ 
KANLib (GRBF)     & $0.5476 \pm 0.0071$ & $0.7772 \pm 0.0058$ & $4050$             & $45.18  \pm 0.57$   \\ 
PyKAN             & $0.5410 \pm 0.0049$ & $0.7825 \pm 0.0039$ & $4050$             & $205.78 \pm 2.42$   \\ 
EfficientKAN      & $0.5453 \pm 0.0059$ & $0.7790 \pm 0.0048$ & $4050$             & $142.14 \pm 1.91$   \\ 
KANLib (B-spline) & $0.5399 \pm 0.0068$ & $0.7834 \pm 0.0054$ & $4050$             & $140.73 \pm 1.63$   \\ 
\hline
\end{tabular}
\label{tbl:baseline}
\end{table*}

\subsection{Computational Efficiency}

In addition to predictive performance, we evaluate the computational efficiency of the different KAN implementations by comparing the number of trainable parameters and the inference time on CPU (AMD Ryzen 9 9950X). Inference time is measured by averaging the time required to process $1000$ independently sampled inputs.

As shown in Table~\ref{tbl:baseline}, all evaluated models except FastKAN contain $4050$ trainable parameters. These parameters include spline coefficients $c_i$, residual branch weights $w_b$, and additional spline weights $w_s$. FastKAN contains fewer parameters because it omits the additional spline weighting mechanism implemented in PyKAN, EfficientKAN, and KANLib.

The measured inference times reveal several important trends. First, PyKAN is the slowest implementation with an inference time of $205.78 \pm 2.42$ ms. EfficientKAN and KANLib (B-spline) substantially improve computational efficiency, reducing inference time by approximately $30.9\%$ and $31.6\%$, respectively, compared to PyKAN. The close match between the inference times of EfficientKAN and KANLib (B-spline) is expected, because the KANLib implementation incorporates several optimizations introduced by EfficientKAN.

Replacing B-splines with Gaussian radial basis functions leads to an additional reduction in inference time. KANLib (GRBF) achieves an inference time of $45.18 \pm 0.57$ ms, corresponding to an improvement of approximately $78.0\%$ compared to PyKAN. FastKAN achieves the fastest execution with $41.15 \pm 0.52$ ms and therefore improves inference speed by approximately $80.0\%$ relative to PyKAN.

\subsection{KAN Architecture Exploration}

As discussed in Section~\ref{sec:kanlib}, one of the primary goals of KANLib is to provide a flexible research framework for exploring different KAN architectures. Consequently, the framework must support modifications of the traditional KAN layer structure defined in Equation~\ref{eq:learnable_function}.

Table~\ref{tbl:kanlib} demonstrates that KANLib supports several architectural variations beyond the standard KAN formulation. The \textit{Default} configuration corresponds to Equation~\ref{eq:learnable_function} and includes both the residual branch $w_b b(x)$ and the spline weight $w_s$. The \textit{Plain} configuration represents a minimal KAN layer in which both the residual branch and spline weights are removed, leaving only the spline computations. All experiments use the same two-layer architecture and training setup introduced in Section~\ref{sec:experimental_setup}.

\begin{table*}
\centering
\caption{Performance of different KANLib architecture configurations on the California Housing dataset across $20$ training runs. Results are reported as mean $\pm$ standard deviation for validation RMSE, $R^2$, and inference time.}
\begin{tabular}{|l|l|rr|rr|}
\hline
Basis Function               & Configuration    & RMSE                & $R^2$               & $\#_\text{params}$ & Inference Time [ms] \\
\hline
\multirow[c]{4}{*}{B-spline} & Default          & $0.5399 \pm 0.0068$ & $0.7834 \pm 0.0054$ & $4050$             & $139.45 \pm 1.76$ \\
                             & No Residual      & $0.5353 \pm 0.0050$ & $0.7871 \pm 0.0040$ & $3780$             & $131.43 \pm 1.95$ \\
                             & No Spline Weight & $0.5495 \pm 0.0055$ & $0.7757 \pm 0.0045$ & $3780$             & $135.67 \pm 2.33$ \\
                             & Plain            & $0.5451 \pm 0.0041$ & $0.7793 \pm 0.0033$ & $3510$             & $127.68 \pm 1.54$ \\
\hline
\multirow[c]{4}{*}{GRBF}     & Default          & $0.5476 \pm 0.0071$ & $0.7772 \pm 0.0058$ & $4050$             & $44.49  \pm 2.00$ \\
                             & No Residual      & $0.5449 \pm 0.0052$ & $0.7794 \pm 0.0042$ & $3780$             & $35.73  \pm 0.24$ \\
                             & No Spline Weight & $0.5576 \pm 0.0043$ & $0.7690 \pm 0.0035$ & $3780$             & $39.41  \pm 0.30$ \\
                             & Plain            & $0.5565 \pm 0.0044$ & $0.7699 \pm 0.0036$ & $3510$             & $30.01  \pm 0.23$ \\
\hline
\end{tabular}
\label{tbl:kanlib}
\end{table*}

Across both basis function types, the \textit{Default} and \textit{No Residual} configurations achieve the strongest predictive performance. However, the overall differences between configurations remain comparatively small, indicating that all considered variants represent viable architectural alternatives.

More substantial differences can be observed in terms of parameter count and inference time. The \textit{Default} configuration contains $4050$ trainable parameters, while removing either the residual branch or the spline weights reduces the parameter count to $3780$. The \textit{Plain} configuration further reduces the parameter count to $3510$.

The inference time measurements reveal that removing the residual branch produces a larger speedup than removing the spline weights. Relative to the \textit{Default} configuration, removing the residual branch reduces inference time by approximately $5.8\%$ for B-spline-based models and $19.7\%$ for GRBF-based models. Removing the spline weights reduces inference time by approximately $2.7\%$ and $11.4\%$, respectively.

These observations are consistent with the underlying computational structure of the layer implementations. Spline weights represent a comparatively inexpensive operation because they can be folded into the spline coefficients during the forward pass \cite{blealtan_blealtanefficient-kan_2025,hoever_julianhoeverkanlib_2026}. In contrast, the residual branch introduces an additional SiLU activation and a matrix multiplication proportional to the input and output dimensions of the layer.

Overall, our experiments demonstrate that KANLib enables systematic exploration of alternative KAN architectures while maintaining competitive predictive performance. Depending on the selected configuration, inference time can be reduced by up to approximately $8.4\%$ for B-spline-based models and $32.5\%$ for GRBF-based models in the considered experimental setting.

\section{Conclusion}\label{sec:conclusion}

This paper introduced KANLib, a modular and extensible framework for Kolmogorov-Arnold Networks (KANs). KANLib provides a flexible foundation for future research on KAN architectures, basis functions, and optimization strategies.

In contrast to existing implementations such as PyKAN \cite{liu_kan_2024,liu_kan_2025,liu_kindxiaomingpykan_2025}, EfficientKAN \cite{blealtan_blealtanefficient-kan_2025}, and FastKAN \cite{li_kolmogorov-arnold_2024,li_ziyaolifast-kan_2025}, which are typically designed around a single type of basis function and a fixed architectural design, KANLib was developed with a strong focus on research-oriented extensibility and flexibility.

The evaluation shows that KANLib reproduces the predictive behavior of established KAN implementations while maintaining competitive computational performance. Furthermore, the framework supports architectural variations beyond the standard KAN formulation with only minor impacts on predictive quality.

Future work focuses on supporting KAN-based 1D convolutions, an important next step toward applying KANs to time-dependent sensor data such as ECG, EEG, and audio signals. In addition, we want to investigate the continual learning abilities of KANs promised by Liu et al. \cite{liu_kan_2025}, an interesting direction for incremental on-device learning.

\section*{Acknowledgment}
The authors acknowledge using AI tools (ChatGPT, Gemini, and DeepL) for brainstorming, proofreading, and structuring the work.
The authors acknowledge the financial support by the German Federal Ministry for Economic Affairs and Energy (BMWE) and by the Ministry of Economic Affairs, Industry, Climate Action and Energy of the State of North Rhine-Westphalia (MWIKE NRW) in the "5-Standorte Programm", joint project ZaKI.D, funding number: 11-09862.

\bibliographystyle{IEEEtran}
\bibliography{references}

\end{document}